\pdfoutput=1
\documentclass[11pt]{article}

\usepackage{EACL2023}

\usepackage{times}
\usepackage{latexsym}

\usepackage[T1]{fontenc}
\usepackage[utf8]{inputenc}

\usepackage{microtype}
\usepackage{multirow}
\usepackage{graphicx}
\usepackage{lipsum}
\usepackage{subcaption}
\usepackage{float}

\usepackage{inconsolata}

\title{Saved You A Click: Automatically Answering Clickbait Titles}

\author{Oliver Johnson\thanks{equal contribution} \quad Beicheng Lou\footnotemark[1] \quad Janet Zhong\footnotemark[1] \quad Andrey Kurenkov \\
Stanford University \\ 
\texttt{johnsono} \quad \texttt{beichenglou} \quad \texttt{
janetzh} \quad \texttt{andreyk} \\
\texttt{@stanford.edu}
        }

\begin{document}
\maketitle
\begin{abstract}
Often clickbait articles have a title that is phrased as a question or vague teaser that entices the user to click on the link and read the article to find the explanation. We developed a system that will automatically find the answer or explanation of the clickbait hook from the website text so that the user does not need to read through the text themselves. We fine-tune an extractive question and answering model (RoBERTa) and an abstractive one (T5), using data scraped from the 'StopClickbait' Facebook pages and Reddit's 'SavedYouAClick' subforum. We find that both extractive and abstractive models improve significantly after finetuning. We find that the extractive model performs slightly better according to ROUGE scores, while the abstractive one has a slight edge in terms of BERTscores. 
\end{abstract}

\section{Introduction}
% The introduction explains the problem, why it's difficult, interesting, or important, how and why current methods succeed/fail at the problem, and explains the key ideas of your approach and results. Though an introduction covers similar material as an abstract, the introduction gives more space for motivation, detail, references to existing work, and to capture the reader's interest.
% filtering out clickbait does not relieve the anxiety caused by not knowing the answer to the clickbait teaser. It would be useful to have a way to automatically find the answer to the clickbait without having to read the whole article.

News articles often use clickbait titles or teasers (cliff-hangers, titles with insufficient information, sensationalized headlines) in order to attract more page views. These titles can be misleading or facetious, which typically wastes readers' time, even though the clickbait hook could be answered very succinctly. While natural language processing tools exist for clickbait detection~\cite{zuhroh2020clickbait}, there has been little study on \textit{automatically answering} clickbait hooks. The interest in such a problem is apparent - a Reddit subforum called `SavedYouAClick' with 1.8 million users is a forum where users post manually generated answers to clickbait hooks (after having read the article). A similar internet organization called `StopClickbait' has a Facebook page with 200,000 followers. We show that clickbait spoiling can be solved as both an extractive and abstractive question and answering (QA) task, where a QA model takes in article-title pairs and outputs an answer that directly solves the clickbait teaser. 
% In other words, we seek to automate the popular `SavedYouAClick' Reddit subforum or `StopClickbait' Facebook page.

In this report, we scrape a training data set from both the `SavedYouAClick' Reddit subforum ($n=2538$) as well `StopClickbait' Facebook pages ($n=1287$). Each data point consists of (c, q, a) where c is the context (the clickbait article), q is the question input to our QA model (the clickbait title) and a is the answer (the Reddit or Facebook user-generated answer to the clickbait hook). We fine-tune both an extractive model (RoBERTa~\cite{RoBERTa} pretrained with SQuAD2.0 or NewsQA) and an abstractive model (T5 abs-qa~\cite{roberts2020much}) on the combined dataset. We evaluate and compare their performances both qualitatively and quantitatively.

\section{Related Work}
Question and answering (QA) is separated into closed-domain and open-domain QA. As our model aims to find the answer within a single provided piece of text (the clickbait article), our problem is a closed-domain QA problem. There are two main approaches to QA:
\begin{itemize}
    \item \textbf{Extractive QA} models extract the answer to a given question from within a provided context~\cite{rajpurkar-etal-2016-squad,rajpurkar2018know,zhu2020question,assem2021qasar,ishigaki2020distant,varanasi2021autoeqa}. In order to train or fine-tune extractive QA models, it requires labeled answer spans of where the answer is contained in the context. Most state-of-the-art results on QA data sets to date are from extractive QA models.
    \item \textbf{Abstractive QA} does not require a labelled answer span. Instead, abstractive models generate new sentences from information or meaning extracted from the corpus. It can output new phrases and sentences that are not in the original source text which makes it more closely related to natural language generation. 
    % The typical approach to handle input questions are to put the context and question together as the input sequence. The model then ouputs a sequence corresponding to the desired answer. 
    It is sometimes referred to as query-based abstractive summarization~\cite{hasselqvist2017query,nema2017diversity} and generative QA (see ERNIE-GEN~\cite{xiao2020ernie}). Significant advances have been made in abstractive QA after Google's T5 model~\cite{raffel2019exploring} was adapted to QA tasks~\cite{roberts2020much}.
    \end{itemize}

Abstractive models can be more powerful than extractive models where the answer is found in many parts throughout the text (such as long-form QA tasks like the ELI5 task~\cite{fan2019eli5}), while sometimes it generates factually incorrect sentences~\cite{kryscinski2019evaluating}. 
A recent work~\cite{hagen2022clickbait} uses extractive QA models to solve for clickbait spoilers based on the Webis Clickbait Spoiling Corpus. Their study does not consider abstractive QA model.

% \subsection{Automatic annotation of answer span for custom data sets in extractive QA}
In order to fine-tune or train an extractive question and answering model, we require a data set that contains (c,q,a,s), where c is the context, q is the question, a is the answer and s is the answer span of the answer within the context. In our data set, we have (c,q,a) where c is the news article, q is the clickbait title and a is the user-generated answer to the clickbait hook. However, we do not have labelled answer spans (s) of where to find the answer within our articles. For most extractive QA data sets, these answer spans are manually annotated by humans (as in NewsQA for example~\cite{trischler2016newsqa}) which can be costly and impractical for large data sets. A few approaches to automatically label answer spans include using exact match, semantic similarity searchers or BERT models such as spanBERT~\cite{joshi2020spanbert}. Ref.~\cite{assem2021qasar} introduces a self-supervised learning framework Qasar, which uses spanBERT and an SBERT-context retriever to locate answer spans. Similarly, in Ref.~\cite{varanasi2021autoeqa}, they propose a model called AutoEQA which automatically annotates answer spans by inserting a QA step in the encoding process and a question generating (QG) step in the decoding process. Another approach is to use distant supervision~\cite{ishigaki2020distant}, which involves training a model on pseudo data that can automatically label answer spans using different heuristics, rules,
and/or external resources. These methods assume (c,q,a) are known, as is the case for our data set. 
% To summarize, the automatic labelling of answer spans for fine-tuning extractive QA models with custom data sets is a nontrivial task. In contrast, fine-tuning or training abstractive QA models only require (c,q,a) data sets, which could be computationally simpler to implement.

% \subsection{Clickbait spoiling}
% A recent work~\cite{hagen2022clickbait} uses extractive QA models to solve for clickbait spoilers based on the Webis Clickbait Spoiling Corpus. Their study does not consider abstractive QA model.

\section{Approach}
% This section details your approach(es) to the problem. 
% For example, this is where you describe the architecture of your neural network(s), and any other key methods or algorithms.

For our clickbait title QA problem, we fine-tuned both an extractive and abstractive QA model. We discuss the process of scraping our data sets in section~\ref{sec:data} which later informs decisions in our approach for the two QA model types.

\subsection{Data}\label{sec:data}
% Describe the dataset(s) you are using (provide references). If it's not already clear, make sure the associated task is clearly described.
% Being precise about the exact form of the input and output can be very useful for readers attempting to understand your work, especially if you've defined your own task.
We identified two pre-existing sources that contained data in the a (c,q,a) format, where c is the context (clickbait news article), q is the question (clickbait title) and a is the answer (manual user-generated post that answers the clickbait hook). These two sources were the Reddit subforum \href{https://www.reddit.com/r/savedyouaclick/}{r/savedyouaclick} and the \href{https://www.facebook.com/StopClickBaitOfficial/}{@StopClickBaitOfficial} Facebook pages. We also include subcategory `StopClickbait,' Facebook pages \href{https://www.facebook.com/SCBEntertainment/}{@SCBEntertainment}, \href{https://www.facebook.com/SCBGaming/}{@SCBGaming}, \href{https://www.facebook.com/SCBLifestyle/}{@SCBLifestyle}, \href{https://www.facebook.com/SCBScience/}{@SCBScience}, \href{https://www.facebook.com/SCBSports/}{@SCBSports} and \href{https://www.facebook.com/SCBNews/}{@SCBNews}. The scraped data from the Reddit and Facebook data sets are both split into training/validation/test set at the ratio of 8/1/1.

\subsubsection{Reddit `SavedYouAClick' $n=2538$ data points}
    We scraped around 25,000 posts from the 'SavedYouAClick' Reddit subforum. After cleaning the dataset and filtering for those where the answer span could be automatically labelled (and manually verified), 2535 posts remained. As the Reddit posts were contributed by many different users, there were often inconsistencies in tone. The user-answers to the clickbait hooks also sometimes included user opinions (`waste of time article') or superfluous details (`saved 31 clicks') that were less relevant to our clickbait QA task. Another potential issue with the Reddit data sets were that some users also posted to solicit a funny or humorous opinion, as opposed to accurately answering the clickbait hook. Fine-tuning our QA model with such data may allow our model to learn these humorous elements (see section~\ref{sec:sarcastic}), at the cost of more noise in the data. 
    %%% limitations
    % There is also variability in how effectively the users answer the clickbait title. After manual inspection, we noticed that some user-generated answers were not good summaries or responses to the clickbait titles and training our models on such data may also introduce some noise.

\subsubsection{Facebook `StopClickbait' $n=1287$ data points}
   We scraped the StopClickbait Facebook pages and obtained a dataset of $n=1287$ (article, title, user-answer to clickbait hook) datapoints after cleaning. As opposed to the Reddit data set, there was more uniformity in the style and tone of this Facebook data set, which may make it more effective for fine-tuning a QA model.

\subsubsection{Fine-tune extractive QA model}\label{sec:extmodel}
The steps for fine-tuning an extractive model given (c,q,a) data has two parts:
\begin{enumerate}
    \item\textbf{ Locate answer span within context} \\
    Our training data set needs to be in a (c,q,a,s) format, but our scraped data set only contains (c, q, a). In order to locate the answer spans (s), we used string similarity search and manual inspection. Sometimes the answer to the clickbait hook (the question/ title) could not be found with the article (the context). An example is when a holistic summary of the article is the answer to the clickbait hook or when the answer appears multiple times in the text. We deleted these data points from our data set. 
    % In future, it would be possible to include multiple answer spans for each clickbait title, rather than just one.
    \item \textbf{Fine-tune the extractive model}\\
    For the extractive QA model, we used two variations of RoBERTa \cite{RoBERTa} which are pretrained on either SQUAD 2.0 and NewsQA\footnote{These are the HuggingFace models \texttt{deepset/roberta-base-squad2} and \texttt{tli8hf/unqover-roberta-base-newsqa} respectively.}. The baseline for our extractive model is the pretrained models without finetuning.
\end{enumerate}

\subsubsection{Fine-tune abstractive QA model}\label{sec:absmodel}
Unlike the extractive model, fine-tuning an abstractive model does not require a (c, q, a, s) format for the data set, so little preprocessing is needed. We start with a variation of the T5 model~\cite{raffel2019exploring,roberts2020much}, pretrained for abstractive QA\footnote{This is the \texttt{tuner007/t5\textunderscore abs\textunderscore qa} model in HuggingFace.}. We fine-tuned this further with our Facebook data, our Reddit data, our Facebook and Reddit data combined and the data set used in Ref.~\cite{hagen2022clickbait}. Of these, we picked the best performing fine-tuned model according to our evaluation metrics. The input sequence in such a model is a combination of the context (the news article) and the query (the click-bait title). The model will then perform abstractive QA by outputting a query-based abstractive summary of the text. The baseline for our abstractive model is the T5 model without finetuning.
%The input is of the form \lstinline{"[context] <question for context: [query]"}. 

\section{Experiments}
\subsection{Experimental details}\label{sec:expdetails}
For the extractive model, we fine-tuned with AdamW optimization for 5 epochs, with a learning rate of $3 \times 10^{-5}$, a batch size of 16.
% The extractive model without fine-tuning already performed reasonably well (and better than the abstractive model without fine-tuning, as seen by the BERTscores in Table~\ref{tab:my-table}.
For the abstractive model, we fine-tuned with AdamW optimization for 20 epochs, with a learning rate of $5 \times 10^{-5}$, a batch size of 2.
The code is available online\footnote{The code is at \href{https://github.com/janetzhong/Saved-You-A-Click-CS224N}{github.com/janetzhong/Saved-You-A-Click-CS224N}.}

% This section contains the following.
%We describe our methods for evaluation in section~\ref{sec:eval}, experimental details in section~\ref{sec:expdetails} and our quantitative results in section~\ref{sec:results}
\subsection{Evaluation method}\label{sec:eval}
% Describe the evaluation metric(s) you use, plus any other details necessary to understand your evaluation.
% Some projects will have clear metrics from prior work on given datasets, but we realize that other projects will define their own metrics.
% If you're defining your own metrics, be clear as to what you're hoping to measure with each evaluation method (whether quantitative or qualitative, automatic or human-defined!), and how it's defined.
We evaluate our fine-tuned QA models by comparing the output of our QA model to the manually generated answers to the clickbait hook by the Reddit and Facebook users. In this case, these Reddit or Facebook answers are the ground truth answers Although these answers are typically quite good (based on manual inspection), in some cases these user-generated answers do not perfectly answer the clickbait article, and this could influence our evaluation metric calculations. Among many evaluation metrics~\cite{fabbri2021summeval}, we choose ROUGE~\cite{lin2004rouge} and BERTscore~\cite{zhang2019bertscore} for the reasons listed below.

\begin{itemize}
    \item \textbf{ROUGE} \\
    Since our QA model can be quite similar to a summarization task, we use ROUGE scores to evaluate our models as this is one of the most widely used metrics for summarization. Each ROUGE score has a recall, precision and F1-measure. ROUGE-1 and ROUGE-2 measure the unigram and bigram overlap respectively, whereas ROUGE-L measures the longest common subsequence between the model output and the reference. Despite being widely used, ROGUE has limitations when an output and reference are paraphrased. Semantitcally correct but paraphrased answers are penalized via ROUGE, which could result in deceptively low ROUGE scores for some cases in our dataset. \\
    %The recall gives an indication of how many words or n-grams of the reference answer is in the system generated answer. However, this disproportionately favours long system generated answers. Thus, the precision measures how much of the system generated answer was relevant in comparison to the reference answer. The F1-measure combines the information from the precision and the recall. 
    \vspace{-3ex}
    \item \textbf{BERTscore} \\
    In order to circumvent this issue of poor evaluation of paraphrased reference and output answers, we also use BERTscore which measure semantic similarity instead of exact word or n-gram overlap. It does this by using contextual word embeddings from BERT and calculating the cosine similarity of the output and the reference. The BERTScore also computes a precision, recall, and F1 measure. BERTscores have been shown to have better correlation with human judgement than ROUGE scores for natural language generation tasks~\cite{zhang2019bertscore}.
\end{itemize}

\subsection{Results}\label{sec:results}
% Report the quantitative results that you have found so far. Use a table or plot to compare results and compare against baselines.
% \begin{itemize}
%     \item If you're a default project team, you should \textbf{report the F1 and EM scores you obtained on the test leaderboard} in this section. Make it clear whether you are on the non-PCE or PCE leaderboard. You can also report dev set results if you like. 
%     \item Comment on your quantitative results. Are they what you expected? Better than you expected? Worse than you expected? Why do you think that is? What does that tell you about your approach?
% \end{itemize}
Our evaluation metrics are calculated for all models using the test data split from the combined data set. We list the ROUGE and BERTscores for the extractive models and abstractive QA models before and after fine-tuning in Table~\ref{tab:my-table}, where e.g. 'RoBERTa NewsQA Reddit' stands for RoBERTa pretrained with NewsQA fine-tuned with the Reddit data set.
% 'T5 Facebook Reddit' is the T5 model fine-tuned with our Facebook and Reddit data set, 'RoBERTa SQuAD' is RoBERTa pretrained with SQuAD 2.0, 'RoBERTa SQuAD Reddit' is RoBERTa pretrained with SQuAD 2.0 fine-tuned with the Reddit data set, 'RoBERTa NewsQA' is RoBERTa pretrained with NewsQA, 'RoBERTa NewsQA Reddit' is RoBERTa pretrained with NewsQA fine-tuned with the Reddit data set.
\begin{table*}[!hbt]
\resizebox{\textwidth}{!}{%
\begin{tabular}{ll|llllllllllll|}
\cline{3-14}
 &
   &
  \multicolumn{3}{l|}{ROUGE-1} &
  \multicolumn{3}{l|}{ROUGE-2} &
  \multicolumn{3}{l|}{ROUGE-L} &
  \multicolumn{3}{l|}{BERTscore} \\ \cline{3-14} 
 &
   &
  \multicolumn{1}{l|}{P} &
  \multicolumn{1}{l|}{R} &
  \multicolumn{1}{l|}{F} &
  \multicolumn{1}{l|}{P} &
  \multicolumn{1}{l|}{R} &
  \multicolumn{1}{l|}{F} &
  \multicolumn{1}{l|}{P} &
  \multicolumn{1}{l|}{R} &
  \multicolumn{1}{l|}{F} &
  \multicolumn{1}{l|}{P} &
  \multicolumn{1}{l|}{R} &
  F \\ \hline
\multicolumn{1}{|l|}{\multirow{2}{*}{abs}} &
  T5 &
  13.39 &
  16.52 &
  12.23 &
  5.47 &
  7.49 &
  5.28 &
  12.63 &
  16.02 &
  11.70 &
  86.07 &
  86.21 &
  86.10 \\
\multicolumn{1}{|l|}{} &
  T5 Facebook Reddit &
  48.20 &
  44.70 &
  41.85 &
  32.72 &
  30.04 &
  28.48 &
  47.57 &
  44.33 &
  41.46 &
  90.34 &
  \textbf{90.98} &
  \textbf{90.61} \\ \hline
\multicolumn{1}{|l|}{\multirow{4}{*}{ext}} &
  RoBERTa SQuAD &
  37.15 &
  34.45 &
  31.65 &
  25.05 &
  22.77 &
  21.75 &
  36.42 &
  33.84 &
  31.09 &
  88.74 &
  88.54 &
  88.59 \\
\multicolumn{1}{|l|}{} &
  RoBERTa SQuAD Reddit &
  47.20 &
  45.24 &
  41.17 &
  34.34 &
  32.26 &
  30.34 &
  46.85 &
  44.97 &
  40.91 &
  90.19 &
  90.26 &
  90.18 \\ \cline{2-14} 
\multicolumn{1}{|l|}{} &
  RoBERTa NewsQA &
  27.16 &
  23.82 &
  22.13 &
  15.63 &
  14.49 &
  13.25 &
  26.43 &
  23.28 &
  21.57 &
  87.65 &
  87.60 &
  87.57 \\
\multicolumn{1}{|l|}{} &
  RoBERTa NewsQA Reddit &
  \textbf{51.58} &
  \textbf{46.50} &
  \textbf{43.72} &
  \textbf{37.46} &
  \textbf{32.33} &
  \textbf{31.42} &
  \textbf{51.03} &
  \textbf{46.21} &
  \textbf{43.41} &
  90.41 &
  90.72 &
  90.52 \\ \hline
\end{tabular}%
}
\vspace{5 pt}
\caption{ ROUGE and BERTscores for our abstractive (abs) and extractive (ext) QA models before and after fine-tuning. Precision (P), Recall (R) and F1 (F) scores are shown.}
\label{tab:my-table}
\end{table*}
We see that there are improvements in the ROUGE and BERTscores in all QA models after finetuning with our scraped Facebook or Reddit data sets. Based on F1 BERTscores, we see that the extractive QA models performed better without fine-tuning in comparison to the abstractive QA models. All models improved after fine-tuning, but the abstractive QA model improved by a greater margin than the improvement in the extractive QA models. 
After fine-tuning, the ROUGE scores and BERTscores for the extractive and abstractive models were comparable.
% , with the RoBERTa pretrained with NewsQA, fine-tuned with the Reddit data set performing the best in ROUGE scores and the T5 model fine-tuned with both the Reddit and Facebook data sets performing the best in BERTscores. 
The abstractive model can generate new terms or paraphrases that will not be captured by the ROUGE scores, so it is reasonable that the BERTscores are higher for an abstractive model while the ROUGE scores are higher for an extractive model. 
% We note that while the F1 BERTscore is slightly higher for the abstractive QA model, the difference is not hugely substantial. One should test on a larger data set before arriving at a conclusion that one model outperforms the other, particularly as quantitative evaluation metrics have limitations. 

% \subsection{Clickbait title classifier}
% We also attempted to use a clickbait title classifier that classifies when the clickbait title is more like an inferred (or actual) question or a teaser statement. This was done to see whether the extractive or abstractive performed better at a specific question type. If so, it would be useful to include a classifier that picks whether to use the extractive or abstractive model for a certain question type. However, our test data set is too small to draw conclusions from this classification task, so we list our observations in Appendix~\ref{sec:classifier}.

%what can precision and recall tell us?
% what does the rouge1 rouge2 and rougeL tell us specifically?

\section{Analysis}\label{sec:analysis}
% Your report should include \textit{qualitative evaluation}. That is, try to understand your system (e.g. how it works, when it succeeds and when it fails) by inspecting key characteristics or outputs of your model.

% Please add the following required packages to your document preamble:
% \usepackage{graphicx}
% Please add the following required packages to your document preamble:
% \usepackage{graphicx}

Upon manual inspection of our fine-tuned QA models on our test Facebook data set (see Appendix~\ref{exampeltables}), we find that the QA models are all very effective at answering these clickbait hooks. In most cases, both extractive and abstractive models answer correctly with the same meaning as the Reddit or Facebook ground truth (Table~\ref{tab:success}). These examples suggest that our clickbait answering QA models work as intended. While both models perform well, we also note that the abstractive model appears to answer more fluently with correct capitalization and grammar in comparison to the extractive model. Table~\ref{tab:success} provides one representative example of this difference. 
However, there are also cases where the abstractive model is correct but the extractive is incorrect (Table~\ref{tab:abs}) and vice versa (Table~\ref{tab:ext}). Extractive models typically struggle when the clickbait answer cannot be found within a short answer span, as seen in the example in Table~\ref{tab:abs}. The cases where abstractive answers are weaker tend to be when they are paraphrased to give an incorrect meaning. 
In Table~\ref{tab:abs}, `shooting for the moon,' from the article has been paraphrased to `to get to the moon' which no longer makes sense. As extractive models do not use language generation, they do not have this issue. As expected, we see examples where our QA models can output different words that both have the same meaning (see `xylitol' and `artificial sweetener' in Table~\ref{tab:rougefail}). In such cases, the QA model output is successful for either answer, but they would give different ROUGE scores, which shows the limitation of ROUGE as a metric. Cases where both the extractive and abstractive models fail is when the clickbait title is very vague, and is not an obvious inferred question (as in Table~\ref{tab:imperfectdata}). Table~\ref{tab:imperfectdata} also gives an example of imperfect training data as the Reddit and Facebook ground truths sometimes have superfluous details added by the user that is not related to our clickbait QA goal. Another case where both extractive and abstractive models struggle is when the clickbait title is a list, such as `10 reasons why..' or `Top 5 things to do when...'. An example of this is Table~\ref{tab:list}. As the answer is multispan across the text, both QA models struggle. 
%%% limitations
% In practice, such clickbait titles have obvious linguistic markers and could be filtered out by a classifier. We could then feed these list titles to a QA model specifically intended for generating list-like answers, which would improve the performance of QA on these tasks. 
% In terms of quantitative metrics, an extractive model (RoBERTa pretrained with NewsQA fine-tuned with the Reddit data set) performed slightly better according to ROUGE scores, while an abstractive model (T5 fine-tuned with the Reddit and Facebook data sets) had a slight edge in terms of BERTscores. This is different from the common belief that extractive models tend to be more accurate, largely because the task is abstractive by nature and using excerpts as answer is prone to be lengthier.
% Overall, our clickbait QA models work better than expected and give satisfying answers to the clickbait hooks for most cases. \\

\section{Conclusion}
In summary, we applied natural language processing to answering clickbait titles, by finetuning an abstractive and an extractive summarization model respectively on scraped data. We analyze both qualitatively with examples and quantitatively with ROUGE and BERTscores. Both models perform well after finetuning, with respective pros and cons.

%%% limitations
% One potential improvement is to scrape more data or apply data augmentation techniques. More reference answers could be gathered and human annotation could be utilized to locate answer spans more accurately. One could also improve the quality of answer span detection for training the extractive model, e.g. by using a combination of natural language processing text retrieval techniques or distant supervision. On the other hand, one could incorporate classification so as to decide which model to use and get the best of both worlds.

\section*{Limitations}
There is some variance in the quality of user-generated clickbait answers, which adds noise to the data. It could be mitigated through more coordinated crowdsourcing. More reference answers could be gathered and human annotation could be utilized to locate answer spans more accurately.
Human annotation is also helpful in the evaluation process as the current metrics may not be good at detecting factual errors.
In the model, one could incorporate classification so as to decide which model to use and get the best of both worlds. 

\section*{Ethics Statement}
Our work could lead to tools that help users digest clickbait articles, saving time for users and promoting more succinct online communications.

% \section*{Acknowledgements}

% Entries for the entire Anthology, followed by custom entries
\bibliography{references,anthology,custom}
\bibliographystyle{acl_natbib}

\appendix

\onecolumn
\section{Appendix}

\subsection{Table showing improvement in abstractive QA model after training}
In this section, we give a few representative examples from the validation data set for the abstractive QA model in Table \ref{tab:eg-fb}. For most cases, the model performance improved drastically. We pick a representative example for the rare occurrence of persistent failure in Table \ref{tab:eg-fb1}. The original article is very ambiguous, providing many alternatives, while the user picks the one that they think is most favored. The improved model is already mostly correct, as the full instruction could be to ``tell her directly that she was rude by saying `Just stop' ''. The task is indeed hard due to the many possible answers.

For a large number of samples, the model was able to improve from summarizing part of the article to answering more to the point, as demonstrated in Table \ref{tab:eg-fb2}. Here the full answer is "singing with the belief that no mask was needed". While the original answer arguably summarized the most important part of this message, it is less pertinent in terms of answering the clickbait title. The improved version, on the other hand, is very to the point and even a little terse, as are many Facebook posts.

There are also many cases where the model could not find an answer in the context before finetuning, but the model answers perfectly after finetuning, as shown in Table \ref{tab:eg-fb3}. In this example, the query is not an explicit question, making it ambiguous what is really expected in the answer. The model learns to correctly direct the attention to "word" and returns a succinct answer that is exactly as desired.
    \begin{table}[H]
    \begin{subtable}[h!]{0.95\textwidth}
        \centering
        \begin{tabular}{c|p{95mm}}
          \hline
          Clickbait title & Here’s the Perfect Thing to Say When Someone Is Rude \\
          \hline
          Ground truth from Facebook &  "Just stop.”\\
          \hline 
          Model output before tuning & No, it’s not a joke. \\
          \hline
          Model output after 20 epochs & She was rude. \\
          \hline
       \end{tabular}
       \caption{Failure}
       \label{tab:eg-fb1}
    \end{subtable}
    \hfill
    \begin{subtable}[h!]{0.95\textwidth}
        \centering
        \begin{tabular}{c|p{95mm}}
        \hline
        Clickbait title & Doing this innocent activity indoors can speed up the spread of COVID-19 dramatically \\
          \hline
          Ground truth from Facebook & Singing without a mask on.\\
          \hline 
          Model output before tuning & No mask was needed to protect people from the virus. \\
          \hline
          Model output after 20 epochs & Singing. \\
          \hline
        \end{tabular}
        \caption{Partial success}
        \label{tab:eg-fb2}
     \end{subtable}
     \hfill
     \begin{subtable}[h!]{0.95\textwidth}
        \centering
        \begin{tabular}{c|p{95mm}}
        \hline
        Clickbait title & Adding This 1 Ridiculous Word Will Improve Your Email Response Rate by 8 Percent, Says a New Study \\
          \hline
          Ground truth from Facebook & Cheers.\\
          \hline 
          Model output before tuning & No answer available in context \\
          \hline
          Model output after 20 epochs & Cheers. \\
          \hline
        \end{tabular}
        \caption{Success}
        \label{tab:eg-fb3}
     \end{subtable}

     \caption{Examples of generated summary on validation data from Facebook}
     \label{tab:eg-fb}
\end{table}

\subsection{Sarcastic tone in Reddit dataset}\label{sec:sarcastic}
A detailed example of the generated answer is provided in Table \ref{tab:eg-reddit}, where the model indeed picks up the sarcastic tone.
However, the training dynamics is highly unstable. The model quickly collapses to bad answers that do not make much sense. By carefully screening the data, we noticed that many Reddit posts spend large amount of words criticizing the article, e.g. by saying ``It's all shit ...'' or ``Nowhere did they say ...'', which are not to the point of summarizing at all. The wide existence of these misleading or toxic examples (around 1/5 of the whole dataset) degrades the quality of finetuning. These observations motivated us to switch to the cleaner dataset from Facebook posts, and also clean up the Reddit dataset.

\begin{table}[h!]
      \begin{center}
        \caption{Example of generated summary picking up sarcastic tone from Reddit}
        \label{tab:eg-reddit}
        \begin{tabular}{c|p{95mm}}
    Ground truth from Reddit & Mother was breastfeeding and the Watress paid for one of their pizzas as a show of support. Inspired Mom now trying to raise Breastfeeding Awareness on Social Media... \\
          \hline 
           Model output before tuning & She was a proud mother. \\
           \hline
          Model output after 2 epochs & She was a snob. \\
           \hline
        \end{tabular}\\
      \end{center}
    \end{table}

\section{Examples of fine-tuned extractive and abstractive QA model results}
\label{exampeltables}
The following examples illustrate our qualitative evaluation of our fine-tuned extractive and abstractive QA models, as described in section~\ref{sec:analysis}.

\begin{table}[h!]
      \begin{center}
        \caption{Example successful output from all fine-tuned models.}
        \label{tab:success}
        \begin{tabular}{c|p{89mm}}
                \hline
            Clickbait title & Justin Bieber's Tattoo Artist Reveals The Meaning Behind His Face Tat. \\
           \hline
           Ground truth from Facebook & It's a cross that represents his journey in finding purpose with God.  \\
           \hline
          RoBERTa + NewsQA \\fine-tuned with Reddit data set & represents his journey in finding purpose with God. \\
           \hline
            RoBERTa + SQuAD2.0 \\fine-tuned with Reddit data set & represents his journey in finding purpose with God. \\
           \hline
            T5 model fine-tuned \\with Facebook data set & He finds purpose with God.  \\
           \hline
        \end{tabular}\\
      \end{center}
    \end{table}

\begin{table}[h!]
      \begin{center}
\caption{Example of abstractive QA model succeeding while extractive QA model fails.}
\label{tab:abs}
        \begin{tabular}{c|p{89mm}}
        \hline
            Clickbait title & “My Little Brother Found Out I’m Not His Fully Sister, And I Came Home To This… I’m Crying” \\
           \hline
           Ground truth from Facebook & It's a cross that represents his journey in finding purpose with God.  \\
           \hline
          RoBERTa + NewsQA \\fine-tuned with Reddit data set & a half-sibling? \\
           \hline
            RoBERTa + SQuAD2.0 \\fine-tuned with Reddit data set & his beloved sister is not actually a ‘fully sister                               \\
           \hline
            T5 model fine-tuned \\with Facebook data set & He wrote her a touching letter and gave her snacks.  \\
           \hline
        \end{tabular}\\
      \end{center}
    \end{table}

\begin{table}[h!]
      \begin{center}
\caption{Example of extractive QA model succeeding while abstractive QA model fails.}
\label{tab:ext}
        \begin{tabular}{c|p{89mm}}
        \hline
            Clickbait title & A psychologist reveals the biggest reason people don’t achieve their goals  \\
           \hline
           Ground truth from Facebook & They focus on the outcome and not the process.  \\
           \hline
          RoBERTa + NewsQA \\fine-tuned with Reddit data set & they focus only on the outcome, not the process. \\
           \hline
            RoBERTa + SQuAD2.0 \\fine-tuned with Reddit data set & they focus only on the outcome, not the process.              \\
           \hline
            T5 model fine-tuned \\with Facebook data set & To get to the moon.  \\
           \hline
        \end{tabular}\\
      \end{center}
    \end{table}
    
\begin{table}[h!]
      \begin{center}
\caption{Example of why a ROUGE score may be an imperfect measure.}
\label{tab:rougefail}
        \begin{tabular}{c|p{89mm}}
        \hline
            Clickbait title & FDA Warns Common Food Ingredient Can Be Fatal To Dogs   \\
           \hline
           Ground truth from Facebook & Artificial Sweetener.   \\
           \hline
          RoBERTa + NewsQA \\fine-tuned with Reddit data set & xylitol, an artificial sweetener\\
           \hline
            RoBERTa + SQuAD2.0 \\fine-tuned with Reddit data set & xylitol             \\
           \hline
            T5 model fine-tuned \\with Facebook data set & Artificial sweetener.  \\
           \hline
        \end{tabular}\\
      \end{center}
    \end{table}

\begin{table}[h!]
      \begin{center}
\caption{Example of all models failing for a vague title. Example also illustrates imperfect Facebook and Reddit data due to additional user comments such as `Video in comments. Thank you Brian!'}
\label{tab:imperfectdata}
        \begin{tabular}{c|p{89mm}}
        \hline
            Clickbait title & She Undergoes Surgery For A Tumor Removal But The Doctors Find Something Else Entirely   \\
           \hline
           Ground truth from Facebook & Her undeveloped embryonic twin. Video in comments. Thank you Brian!   \\
           \hline
          RoBERTa + NewsQA \\fine-tuned with Reddit data set & Something Else \\
           \hline
            RoBERTa + SQuAD2.0 \\fine-tuned with Reddit data set & Find Something Else             \\
           \hline
            T5 model fine-tuned \\with Facebook data set & She is a cancer survivor.  \\
           \hline
        \end{tabular}\\
      \end{center}
    \end{table}

\begin{table}[h!]
      \begin{center}
\caption{Example of all models failing to answer list-style clickbait titles.}
\label{tab:list}
        \begin{tabular}{c|p{89mm}}
        \hline
            Clickbait title & Here's What It Takes to Make a Mathematical Genius, According to Science   \\
           \hline
           Ground truth from Facebook & 1.Understand and build language 2.Recognise and exploit hidden structures in data.   \\
           \hline
          RoBERTa + NewsQA \\fine-tuned with Reddit data set & language. \\
           \hline
            RoBERTa + SQuAD2.0 \\fine-tuned with Reddit data set & From language            \\
           \hline
            T5 model fine-tuned \\with Facebook data set & No.    \\
           \hline
        \end{tabular}\\
      \end{center}
    \end{table}

\subsection{Full evaluation table}
\label{app-table}
\begin{table}[H]
\resizebox{\textwidth}{!}{%
\begin{tabular}{ll|llllllllllll|}
\cline{3-14}
 &
   &
  \multicolumn{3}{l|}{ROUGE-1} &
  \multicolumn{3}{l|}{ROUGE-2} &
  \multicolumn{3}{l|}{ROUGE-L} &
  \multicolumn{3}{l|}{BERTscore} \\ \cline{3-14} 
 &
   &
  \multicolumn{1}{l|}{P} &
  \multicolumn{1}{l|}{R} &
  \multicolumn{1}{l|}{F} &
  \multicolumn{1}{l|}{P} &
  \multicolumn{1}{l|}{R} &
  \multicolumn{1}{l|}{F} &
  \multicolumn{1}{l|}{P} &
  \multicolumn{1}{l|}{R} &
  \multicolumn{1}{l|}{F} &
  \multicolumn{1}{l|}{P} &
  \multicolumn{1}{l|}{R} &
  F \\ \hline
\multicolumn{1}{|l|}{\multirow{5}{*}{abs}} &
  T5 &
  13.39 &
  16.52 &
  12.23 &
  5.47 &
  7.49 &
  5.28 &
  12.63 &
  16.02 &
  11.70 &
  86.07 &
  86.21 &
  86.10 \\
\multicolumn{1}{|l|}{} &
  T5 Facebook &
  44.76 &
  38.89 &
  36.98 &
  25.90 &
  23.17 &
  21.83 &
  43.77 &
  38.21 &
  36.27 &
  89.27 &
  91.46 &
  90.30 \\
\multicolumn{1}{|l|}{} &
  T5 Reddit &
  46.86 &
  42.51 &
  39.94 &
  29.89 &
  27.7 &
  25.93 &
  46.20 &
  41.92 &
  39.41 &
  89.71 &
  89.90 &
  89.76 \\
\multicolumn{1}{|l|}{} &
  T5 Facebook Reddit &
  48.20 &
  44.70 &
  41.85 &
  32.72 &
  30.04 &
  28.48 &
  47.57 &
  44.33 &
  41.46 &
  90.34 &
  \textbf{90.98} &
  \textbf{90.61} \\
\multicolumn{1}{|l|}{} &
  T5 Webis &
  45.31 &
  40.70 &
  37.83 &
  30.02 &
  27.26 &
  25.35 &
  44.51 &
  40.24 &
  37.35 &
  89.38 &
  89.39 &
  89.33 \\ \hline
\multicolumn{1}{|l|}{\multirow{4}{*}{ext}} &
  RoBERTa SQuAD &
  37.15 &
  34.45 &
  31.65 &
  25.05 &
  22.77 &
  21.75 &
  36.42 &
  33.84 &
  31.09 &
  88.74 &
  88.54 &
  88.59 \\
\multicolumn{1}{|l|}{} &
  RoBERTa SQuAD Reddit &
  47.20 &
  45.24 &
  41.17 &
  34.34 &
  32.26 &
  30.34 &
  46.85 &
  44.97 &
  40.91 &
  90.19 &
  90.26 &
  90.18 \\ \cline{2-14} 
\multicolumn{1}{|l|}{} &
  RoBERTa NewsQA &
  27.16 &
  23.82 &
  22.13 &
  15.63 &
  14.49 &
  13.25 &
  26.43 &
  23.28 &
  21.57 &
  87.65 &
  87.60 &
  87.57 \\
\multicolumn{1}{|l|}{} &
  RoBERTa NewsQA Reddit &
  \textbf{51.58} &
  \textbf{46.50} &
  \textbf{43.72} &
  \textbf{37.46} &
  \textbf{32.33} &
  \textbf{31.42} &
  \textbf{51.03} &
  \textbf{46.21} &
  \textbf{43.41} &
  \textbf{90.41} &
  90.72 &
  90.52 \\ \hline
\end{tabular}%
}
\vspace{5 pt}
\caption{ ROUGE and BERTscores for our abstractive (abs) and extractive (ext) QA models before and after fine-tuning. Precision (P), Recall (R) and F1 (F) scores are shown.}
\label{fulltable}
\end{table}
In Table~\ref{fulltable}, we include additional experiments of fine-tuning our T5 model to the ones listed in Table~\ref{tab:my-table}. We find that fine-tuning the T5 with the Reddit data set only, improved the ROUGE scores more than fine-tuning with the Facebook data set only, whereas fine-tuning with the Facebook data set only improved the BERTscores than fine-tuning with the Reddit data set only. This makes sense given that our Facebook user-generated answers are more like abstractive summaries, and our Reddit data set only contained data where an answer span could be extracted from the article. The T5 model fine-tuned with both the Reddit and Facebook data set was the best fine-tuned T5 model, which makes sense as it has the largest training data set of the aforementioned models. There are limitations in fine-tuning the extractive models with the Facebook data set, as an answer span cannot always be found for the Facebook data set. We also fine-tuned our T5 model with the $n=5000$ data set of clickbait spoilers in Ref~\cite{hagen2022clickbait}. The quality of this training data is expected to be higher, as they used manual human labels of answer spans which took 560 hours, as opposed to our method of automatically extracting answer spans with string similarity and manual inspection. We found that evaluating the T5 model fine-tuned on the Webis data set on our test data set yields lower ROUGE and BERTscores than the T5 model fine-tuned on our Facebook and Reddit data set. However, we note that this is an unfair comparison. The clickbait 'answer' in our data set is the user-generated Facebook or Reddit answer. However, in Ref.~\cite{hagen2022clickbait}, it is the optimal excerpts from the article. Evaluating on our own test data set where the reference answer is the user-generated answers as opposed to the optimal excerpts disproportionately favours our models. Furthermore, optimal excerpts from the articles tend to be longer than the Facebook or Reddit user-generated answers, so fine-tuning with the Webis data set leads to longer outputs than our models. The F1 ROUGE and BERTscore actually have a non negligible the size of the model output~\cite{sun2019compare}, and this could be another reason why our evaluation metric disproportionately penalizes the T5 model fine-tuned with the Webis data set. Fine-tuning the T5 model with the Webis data also ignores their labelled answer spans, as the abstractive model does not require it, though this is likely one of the stronger aspects of their data set. While human annotation of answer spans for QA data sets is the gold standard for many QA data sets, we note the efficiency of using abstractive models, which in this case saves many hours of annotation with fairly comparable results according to our evaluation metrics. 

\end{document}